\documentclass{article}
\usepackage[final]{nips_2016}
\usepackage[utf8]{inputenc} % allow utf-8 input
\usepackage[T1]{fontenc}    % use 8-bit T1 fonts
\usepackage{hyperref}       % hyperlinks
\usepackage{url}            % simple URL typesetting
\usepackage{booktabs}       % professional-quality tables
\usepackage{amsfonts}       % blackboard math symbols
\usepackage{nicefrac}       % compact symbols for 1/2, etc.
\usepackage{microtype}      % microtypography
\usepackage{graphicx}
\usepackage{xspace}
\usepackage{longtable}

\newcommand{\fmnist}{Fashion-MNIST\xspace}
\newcommand{\repo}{\url{https://github.com/zalandoresearch/fashion-mnist}}
\newcommand{\zraddress}{Mühlenstraße 25, 10243 Berlin}

\title{\fmnist: a Novel Image Dataset for Benchmarking Machine Learning Algorithms}

\author{Han Xiao\\
  Zalando Research\\
  \zraddress\\
  \texttt{han.xiao@zalando.de}\\
  \And
  Kashif Rasul\\
  Zalando Research\\
  \zraddress\\
  \texttt{kashif.rasul@zalando.de}\\
  \And
  Roland Vollgraf\\
  Zalando Research\\
  \zraddress\\
  \texttt{roland.vollgraf@zalando.de}\\
}

\begin{document}

\maketitle

\begin{abstract}
We present \fmnist, a new dataset comprising of $28\times 28$ grayscale images of $70,000$ fashion products from $10$ categories, with $7,000$ images per category. The training set has $60,000$ images and the test set has $10,000$ images. \fmnist is intended to serve as a direct drop-in replacement for the original MNIST dataset for benchmarking machine learning algorithms, as it shares the same image size, data format and the structure of training and testing splits. The dataset is freely available at \repo.
\end{abstract}

\section{Introduction}

The MNIST dataset comprising of 10-class handwritten digits, was first introduced by~\citet{lecun1998gradient} in 1998. At that time one could not have foreseen the stellar rise of deep learning techniques and their performance. Despite the fact that today deep learning can do so much the simple MNIST dataset has become the most widely used testbed in deep learning, surpassing CIFAR-10~\citep{krizhevsky2009learning} and ImageNet~\citep{deng2009imagenet} in its popularity via Google trends\footnote{\url{https://trends.google.com/trends/explore?date=all&q=mnist,CIFAR,ImageNet}}. Despite its simplicity its usage does not seem to be decreasing despite calls for it in the deep learning community.

The reason MNIST is so popular has to do with its size, allowing deep learning researchers to quickly check and prototype their algorithms. This is also complemented by the fact that all machine learning libraries (e.g. scikit-learn) and deep learning frameworks (e.g. Tensorflow, Pytorch) provide helper functions and convenient examples that use MNIST out of the box. 

Our aim with this work is to create a good benchmark dataset which has all the accessibility of MNIST, namely its small size, straightforward encoding and permissive license. We took the approach of sticking to the $10$ classes $70,000$ grayscale images in the size of $28\times 28$ as in the original MNIST. In fact, the only change one needs to use this dataset is to change the URL from where the MNIST dataset is fetched. Moreover, \fmnist poses a more challenging classification task than the simple MNIST digits data, whereas the latter has been trained to accuracies above 99.7\% as reported in~\citet{wan2013regularization,ciregan2012multi}.

We also looked at the EMNIST dataset provided by~\citet{cohen2017emnist}, an extended version of MNIST that extends the number of classes by introducing uppercase and lowercase characters. However, to be able to use it seamlessly one needs to not only extend the deep learning framework's MNIST helpers, but also change the underlying deep neural network to classify these extra classes.

\section{\fmnist Dataset\label{sec:dataset}}
\fmnist is based on the assortment on Zalando's website\footnote{Zalando is the Europe's largest online fashion platform. \url{http://www.zalando.com}}. Every fashion product on Zalando has a set of pictures shot by professional photographers, demonstrating different aspects of the product, i.e. front and back looks, details, looks with model and in an outfit. The original picture has a light-gray background (hexadecimal color: \texttt{\#fdfdfd}) and stored in $762\times 1000$ JPEG format. For efficiently serving different frontend components, the original picture is resampled with multiple resolutions, e.g. large, medium, small, thumbnail and tiny. 

\begin{figure}[h]
  \centering
  \includegraphics[width=1.0\textwidth]{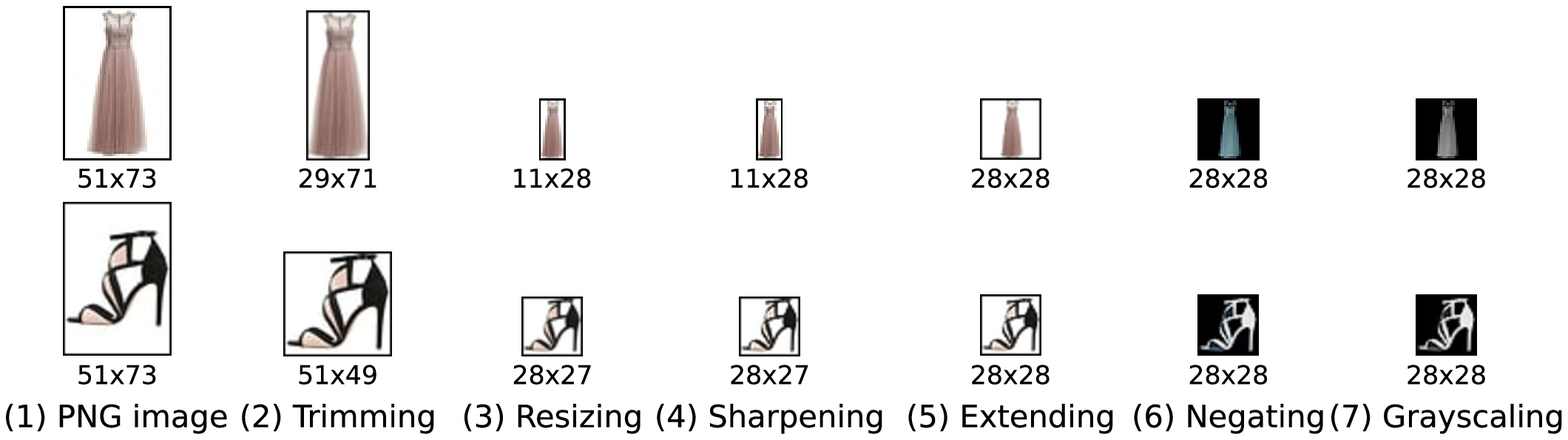}
  \caption{Diagram of the conversion process used to generate \fmnist dataset. Two examples from dress and sandals categories are depicted, respectively. Each column represents a step described in \autoref{sec:dataset}.\label{fig:convert}}
\end{figure}

We use the front look thumbnail images of $70,000$ unique products to build \fmnist. Those products come from different gender groups: men, women, kids and neutral. In particular, white-color products are not included in the dataset as they have low contrast to the background. The thumbnails ($51\times 73$) are then fed into the following conversion pipeline, which is visualized in \autoref{fig:convert}.

\begin{enumerate}
    \item Converting the input to a PNG image.
    \item Trimming any edges that are close to the color of the corner pixels.
    The ``closeness'' is defined by the distance within $5\%$ of the maximum possible intensity in RGB space.
    \item Resizing the longest edge of the image to $28$ by subsampling the pixels, i.e. some rows and columns
    are skipped over.
    \item Sharpening pixels using a Gaussian operator of the radius and standard deviation of $1.0$, with increasing effect near outlines.
    \item Extending the shortest edge to $28$ and put the image to the center of the canvas.
    \item Negating the intensities of the image.
    \item Converting the image to 8-bit grayscale pixels.
\end{enumerate}

\begin{table}[htbp]
  \caption{Files contained in the \fmnist dataset. \label{tbl:files}}
  \centering
  \begin{tabular}{llrr}
    \toprule
    Name     & Description     & \# Examples & Size \\
    \midrule
    train-images-idx3-ubyte.gz     & Training set images & $60,000$ & $25$ MBytes \\
    train-labels-idx1-ubyte.gz & Training set labels & $60,000$ & $140$ Bytes \\
    t10k-images-idx3-ubyte.gz & Test set images  & $10,000$ & $4.2$ MBytes \\
    t10k-labels-idx1-ubyte.gz     & Test set labels & $10,000$ & $92$ Bytes\\
    \bottomrule
  \end{tabular}
\end{table}
For the class labels, we use the silhouette code of the product. The silhouette code is manually labeled by the in-house fashion experts and reviewed by a separate team at Zalando. Each product contains only one silhouette code. \autoref{tbl:example} gives a summary of all class labels in \fmnist with examples for each class.

Finally, the dataset is divided into a training and a test set. The training set receives a randomly-selected $6,000$ examples from each class. Images and labels are stored in the same file format as the MNIST data set, which is designed for storing vectors and multidimensional matrices. The result files are listed in \autoref{tbl:files}. We sort examples by their labels while storing, resulting in smaller label files after compression comparing to the MNIST. It is also easier to retrieve examples with a certain class label. The data shuffling job is therefore left to the algorithm developer.

\begin{table}[htbp]
  \caption{Class names and example images in \fmnist dataset. \label{tbl:example}}
  \centering
  \begin{tabular}{lll}
    \toprule
    Label     & Description     & Examples  \\
    \midrule
$0$ & T-Shirt/Top & \raisebox{-.6\totalheight}{\includegraphics[width=0.72\textwidth]{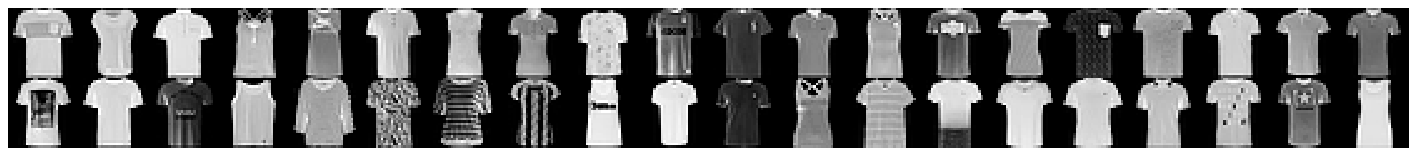}}\\
$1$ & Trouser & \raisebox{-.6\totalheight}{\includegraphics[width=0.72\textwidth]{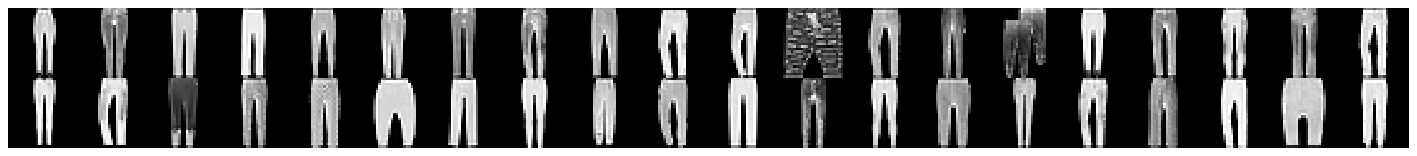}}\\
$2$ & Pullover & \raisebox{-.6\totalheight}{\includegraphics[width=0.72\textwidth]{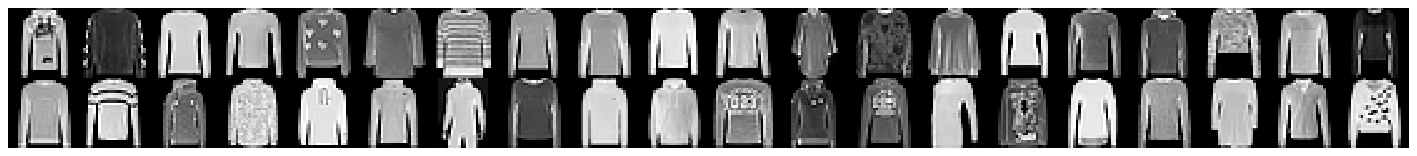}}\\
$3$ & Dress & \raisebox{-.6\totalheight}{\includegraphics[width=0.72\textwidth]{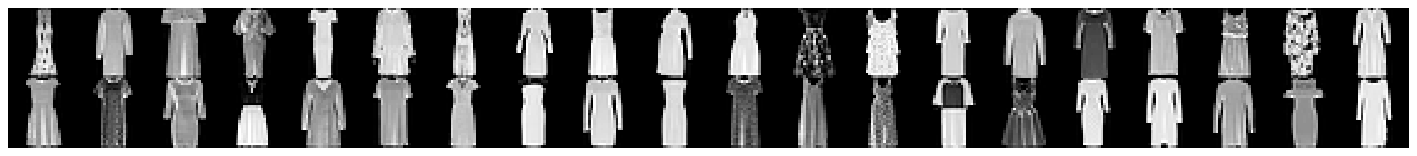}}\\
$4$ & Coat & \raisebox{-.6\totalheight}{\includegraphics[width=0.72\textwidth]{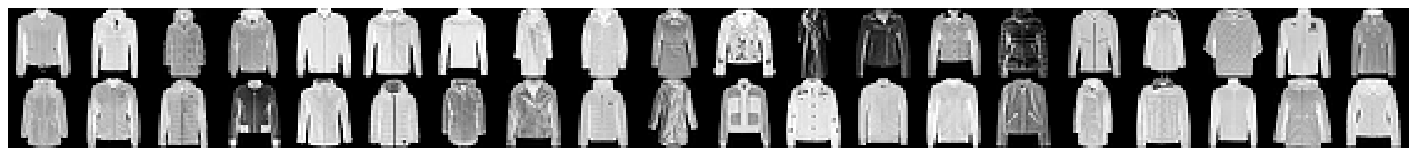}}\\
$5$ & Sandals & \raisebox{-.6\totalheight}{\includegraphics[width=0.72\textwidth]{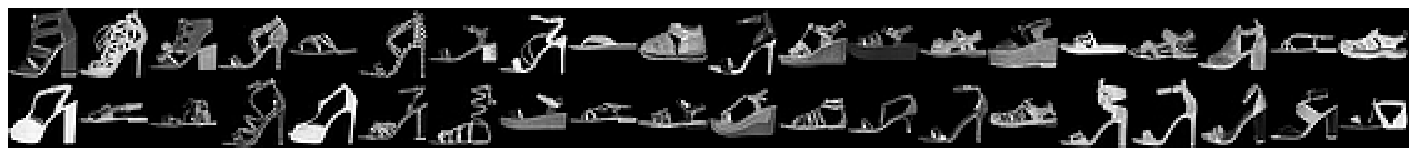}}\\
$6$ & Shirt & \raisebox{-.6\totalheight}{\includegraphics[width=0.72\textwidth]{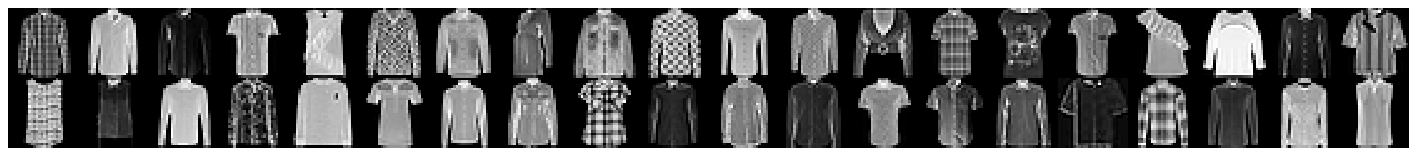}}\\
$7$ & Sneaker & \raisebox{-.6\totalheight}{\includegraphics[width=0.72\textwidth]{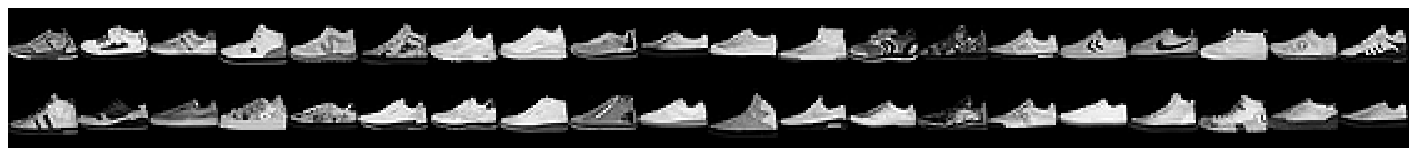}}\\
$8$ & Bag & \raisebox{-.6\totalheight}{\includegraphics[width=0.72\textwidth]{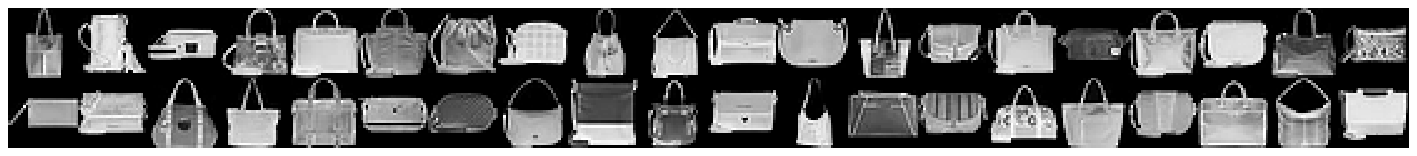}}\\
$9$ & Ankle boots & \raisebox{-.6\totalheight}{\includegraphics[width=0.72\textwidth]{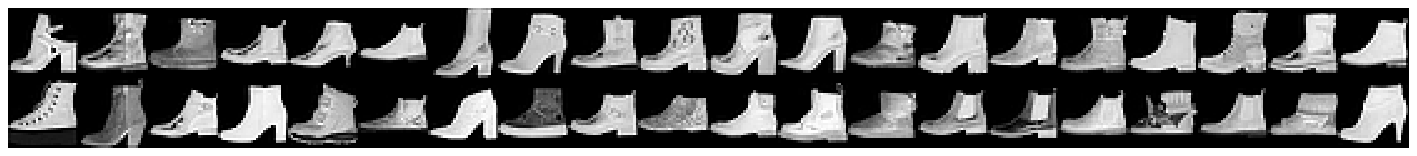}}\\
    \bottomrule
  \end{tabular}
\end{table}

\section{Experiments}
We provide some classification results in \autoref{tbl:benchmark} to form a benchmark on this data set. All algorithms are repeated $5$ times by shuffling the training data and the average accuracy on the test set is reported. The benchmark on the MNIST dataset is also included for a side-by-side comparison. A more comprehensive  table with explanations on the algorithms can be found on \repo.

\begin{longtable}{llrr}
\caption{Benchmark on \fmnist (Fashion) and MNIST.} \label{tbl:benchmark} \\
\toprule
& & \multicolumn{2}{c}{Test Accuracy} \\
\cmidrule{3-4}
\multicolumn{1}{l}{Classifier} &
\multicolumn{1}{l}{Parameter} &
\multicolumn{1}{c}{Fashion} &
\multicolumn{1}{c}{MNIST} \\ 
\midrule
\endfirsthead

\multicolumn{4}{c}%
{{\tablename\ \thetable{} -- continued from previous page}} \\
\toprule
& & \multicolumn{2}{c}{Test Accuracy} \\
\cmidrule{3-4}
\multicolumn{1}{l}{Classifier} &
\multicolumn{1}{l}{Parameter} &
\multicolumn{1}{c}{Fashion} &
\multicolumn{1}{c}{MNIST} \\ 
\midrule
\endhead

\midrule \multicolumn{4}{r}{{Continued on next page}} \\ \bottomrule
\endfoot

\bottomrule
\endlastfoot
DecisionTreeClassifier & \tiny{criterion=\texttt{entropy} max\_depth=$10$ splitter=\texttt{best}} &$0.798$ & $0.873$\\
& \tiny{criterion=\texttt{entropy} max\_depth=$10$ splitter=\texttt{random}} &$0.792$ & $0.861$\\
& \tiny{criterion=\texttt{entropy} max\_depth=$50$ splitter=\texttt{best}} &$0.789$ & $0.886$\\
& \tiny{criterion=\texttt{entropy} max\_depth=$100$ splitter=\texttt{best}} &$0.789$ & $0.886$\\
& \tiny{criterion=\texttt{gini} max\_depth=$10$ splitter=\texttt{best}} &$0.788$ & $0.866$\\
& \tiny{criterion=\texttt{entropy} max\_depth=$50$ splitter=\texttt{random}} &$0.787$ & $0.883$\\
& \tiny{criterion=\texttt{entropy} max\_depth=$100$ splitter=\texttt{random}} &$0.787$ & $0.881$\\
& \tiny{criterion=\texttt{gini} max\_depth=$100$ splitter=\texttt{best}} &$0.785$ & $0.879$\\
& \tiny{criterion=\texttt{gini} max\_depth=$50$ splitter=\texttt{best}} &$0.783$ & $0.877$\\
& \tiny{criterion=\texttt{gini} max\_depth=$10$ splitter=\texttt{random}} &$0.783$ & $0.853$\\
& \tiny{criterion=\texttt{gini} max\_depth=$50$ splitter=\texttt{random}} &$0.779$ & $0.873$\\
& \tiny{criterion=\texttt{gini} max\_depth=$100$ splitter=\texttt{random}} &$0.777$ & $0.875$\\
\midrule
ExtraTreeClassifier & \tiny{criterion=\texttt{gini} max\_depth=$10$ splitter=\texttt{best}} &$0.775$ & $0.806$\\
& \tiny{criterion=\texttt{entropy} max\_depth=$100$ splitter=\texttt{best}} &$0.775$ & $0.847$\\
& \tiny{criterion=\texttt{entropy} max\_depth=$10$ splitter=\texttt{best}} &$0.772$ & $0.810$\\
& \tiny{criterion=\texttt{entropy} max\_depth=$50$ splitter=\texttt{best}} &$0.772$ & $0.847$\\
& \tiny{criterion=\texttt{gini} max\_depth=$100$ splitter=\texttt{best}} &$0.769$ & $0.843$\\
& \tiny{criterion=\texttt{gini} max\_depth=$50$ splitter=\texttt{best}} &$0.768$ & $0.845$\\
& \tiny{criterion=\texttt{entropy} max\_depth=$50$ splitter=\texttt{random}} &$0.752$ & $0.826$\\
& \tiny{criterion=\texttt{entropy} max\_depth=$100$ splitter=\texttt{random}} &$0.752$ & $0.828$\\
& \tiny{criterion=\texttt{gini} max\_depth=$50$ splitter=\texttt{random}} &$0.748$ & $0.824$\\
& \tiny{criterion=\texttt{gini} max\_depth=$100$ splitter=\texttt{random}} &$0.745$ & $0.820$\\
& \tiny{criterion=\texttt{gini} max\_depth=$10$ splitter=\texttt{random}} &$0.739$ & $0.737$\\
& \tiny{criterion=\texttt{entropy} max\_depth=$10$ splitter=\texttt{random}} &$0.737$ & $0.745$\\
\midrule
GaussianNB & \tiny{priors=\texttt{[0.1, 0.1, 0.1, 0.1, 0.1, 0.1, 0.1, 0.1, 0.1, 0.1]}} &$0.511$ & $0.524$\\
\midrule
GradientBoostingClassifier & \tiny{n\_estimators=$100$ loss=\texttt{deviance} max\_depth=$10$} &$0.880$ & $0.969$\\
& \tiny{n\_estimators=$50$ loss=\texttt{deviance} max\_depth=$10$} &$0.872$ & $0.964$\\
& \tiny{n\_estimators=$100$ loss=\texttt{deviance} max\_depth=$3$} &$0.862$ & $0.949$\\
& \tiny{n\_estimators=$10$ loss=\texttt{deviance} max\_depth=$10$} &$0.849$ & $0.933$\\
& \tiny{n\_estimators=$50$ loss=\texttt{deviance} max\_depth=$3$} &$0.840$ & $0.926$\\
& \tiny{n\_estimators=$10$ loss=\texttt{deviance} max\_depth=$50$} &$0.795$ & $0.888$\\
& \tiny{n\_estimators=$10$ loss=\texttt{deviance} max\_depth=$3$} &$0.782$ & $0.846$\\
\midrule
KNeighborsClassifier & \tiny{weights=\texttt{distance} n\_neighbors=$5$ p=$1$} &$0.854$ & $0.959$\\
& \tiny{weights=\texttt{distance} n\_neighbors=$9$ p=$1$} &$0.854$ & $0.955$\\
& \tiny{weights=\texttt{uniform} n\_neighbors=$9$ p=$1$} &$0.853$ & $0.955$\\
& \tiny{weights=\texttt{uniform} n\_neighbors=$5$ p=$1$} &$0.852$ & $0.957$\\
& \tiny{weights=\texttt{distance} n\_neighbors=$5$ p=$2$} &$0.852$ & $0.945$\\
& \tiny{weights=\texttt{distance} n\_neighbors=$9$ p=$2$} &$0.849$ & $0.944$\\
& \tiny{weights=\texttt{uniform} n\_neighbors=$5$ p=$2$} &$0.849$ & $0.944$\\
& \tiny{weights=\texttt{uniform} n\_neighbors=$9$ p=$2$} &$0.847$ & $0.943$\\
& \tiny{weights=\texttt{distance} n\_neighbors=$1$ p=$2$} &$0.839$ & $0.943$\\
& \tiny{weights=\texttt{uniform} n\_neighbors=$1$ p=$2$} &$0.839$ & $0.943$\\
& \tiny{weights=\texttt{uniform} n\_neighbors=$1$ p=$1$} &$0.838$ & $0.955$\\
& \tiny{weights=\texttt{distance} n\_neighbors=$1$ p=$1$} &$0.838$ & $0.955$\\
\midrule
LinearSVC & \tiny{loss=\texttt{hinge} C=$1$ multi\_class=\texttt{ovr} penalty=\texttt{l2}} &$0.836$ & $0.917$\\
& \tiny{loss=\texttt{hinge} C=$1$ multi\_class=\texttt{crammer\_singer} penalty=\texttt{l2}} &$0.835$ & $0.919$\\
& \tiny{loss=\texttt{squared\_hinge} C=$1$ multi\_class=\texttt{crammer\_singer} penalty=\texttt{l2}} &$0.834$ & $0.919$\\
& \tiny{loss=\texttt{squared\_hinge} C=$1$ multi\_class=\texttt{crammer\_singer} penalty=\texttt{l1}} &$0.833$ & $0.919$\\
& \tiny{loss=\texttt{hinge} C=$1$ multi\_class=\texttt{crammer\_singer} penalty=\texttt{l1}} &$0.833$ & $0.919$\\
& \tiny{loss=\texttt{squared\_hinge} C=$1$ multi\_class=\texttt{ovr} penalty=\texttt{l2}} &$0.820$ & $0.912$\\
& \tiny{loss=\texttt{squared\_hinge} C=$10$ multi\_class=\texttt{ovr} penalty=\texttt{l2}} &$0.779$ & $0.885$\\
& \tiny{loss=\texttt{squared\_hinge} C=$100$ multi\_class=\texttt{ovr} penalty=\texttt{l2}} &$0.776$ & $0.873$\\
& \tiny{loss=\texttt{hinge} C=$10$ multi\_class=\texttt{ovr} penalty=\texttt{l2}} &$0.764$ & $0.879$\\
& \tiny{loss=\texttt{hinge} C=$100$ multi\_class=\texttt{ovr} penalty=\texttt{l2}} &$0.758$ & $0.872$\\
& \tiny{loss=\texttt{hinge} C=$10$ multi\_class=\texttt{crammer\_singer} penalty=\texttt{l1}} &$0.751$ & $0.783$\\
& \tiny{loss=\texttt{hinge} C=$10$ multi\_class=\texttt{crammer\_singer} penalty=\texttt{l2}} &$0.749$ & $0.816$\\
& \tiny{loss=\texttt{squared\_hinge} C=$10$ multi\_class=\texttt{crammer\_singer} penalty=\texttt{l2}} &$0.748$ & $0.829$\\
& \tiny{loss=\texttt{squared\_hinge} C=$10$ multi\_class=\texttt{crammer\_singer} penalty=\texttt{l1}} &$0.736$ & $0.829$\\
& \tiny{loss=\texttt{hinge} C=$100$ multi\_class=\texttt{crammer\_singer} penalty=\texttt{l1}} &$0.516$ & $0.759$\\
& \tiny{loss=\texttt{hinge} C=$100$ multi\_class=\texttt{crammer\_singer} penalty=\texttt{l2}} &$0.496$ & $0.753$\\
& \tiny{loss=\texttt{squared\_hinge} C=$100$ multi\_class=\texttt{crammer\_singer} penalty=\texttt{l1}} &$0.492$ & $0.746$\\
& \tiny{loss=\texttt{squared\_hinge} C=$100$ multi\_class=\texttt{crammer\_singer} penalty=\texttt{l2}} &$0.484$ & $0.737$\\
\midrule
LogisticRegression & \tiny{C=$1$ multi\_class=\texttt{ovr} penalty=\texttt{l1}} &$0.842$ & $0.917$\\
& \tiny{C=$1$ multi\_class=\texttt{ovr} penalty=\texttt{l2}} &$0.841$ & $0.917$\\
& \tiny{C=$10$ multi\_class=\texttt{ovr} penalty=\texttt{l2}} &$0.839$ & $0.916$\\
& \tiny{C=$10$ multi\_class=\texttt{ovr} penalty=\texttt{l1}} &$0.839$ & $0.909$\\
& \tiny{C=$100$ multi\_class=\texttt{ovr} penalty=\texttt{l2}} &$0.836$ & $0.916$\\
\midrule
MLPClassifier & \tiny{activation=\texttt{relu} hidden\_layer\_sizes=\texttt{[100]}} &$0.871$ & $0.972$\\
& \tiny{activation=\texttt{relu} hidden\_layer\_sizes=\texttt{[100, 10]}} &$0.870$ & $0.972$\\
& \tiny{activation=\texttt{tanh} hidden\_layer\_sizes=\texttt{[100]}} &$0.868$ & $0.962$\\
& \tiny{activation=\texttt{tanh} hidden\_layer\_sizes=\texttt{[100, 10]}} &$0.863$ & $0.957$\\
& \tiny{activation=\texttt{relu} hidden\_layer\_sizes=\texttt{[10, 10]}} &$0.850$ & $0.936$\\
& \tiny{activation=\texttt{relu} hidden\_layer\_sizes=\texttt{[10]}} &$0.848$ & $0.933$\\
& \tiny{activation=\texttt{tanh} hidden\_layer\_sizes=\texttt{[10, 10]}} &$0.841$ & $0.921$\\
& \tiny{activation=\texttt{tanh} hidden\_layer\_sizes=\texttt{[10]}} &$0.840$ & $0.921$\\
\midrule
PassiveAggressiveClassifier & \tiny{C=$1$} &$0.776$ & $0.877$\\
& \tiny{C=$100$} &$0.775$ & $0.875$\\
& \tiny{C=$10$} &$0.773$ & $0.880$\\
\midrule
Perceptron & \tiny{penalty=\texttt{l1}} &$0.782$ & $0.887$\\
& \tiny{penalty=\texttt{l2}} &$0.754$ & $0.845$\\
& \tiny{penalty=\texttt{elasticnet}} &$0.726$ & $0.845$\\
\midrule
RandomForestClassifier & \tiny{n\_estimators=$100$ criterion=\texttt{entropy} max\_depth=$100$} &$0.873$ & $0.970$\\
& \tiny{n\_estimators=$100$ criterion=\texttt{gini} max\_depth=$100$} &$0.872$ & $0.970$\\
& \tiny{n\_estimators=$50$ criterion=\texttt{entropy} max\_depth=$100$} &$0.872$ & $0.968$\\
& \tiny{n\_estimators=$100$ criterion=\texttt{entropy} max\_depth=$50$} &$0.872$ & $0.969$\\
& \tiny{n\_estimators=$50$ criterion=\texttt{entropy} max\_depth=$50$} &$0.871$ & $0.967$\\
& \tiny{n\_estimators=$100$ criterion=\texttt{gini} max\_depth=$50$} &$0.871$ & $0.971$\\
& \tiny{n\_estimators=$50$ criterion=\texttt{gini} max\_depth=$50$} &$0.870$ & $0.968$\\
& \tiny{n\_estimators=$50$ criterion=\texttt{gini} max\_depth=$100$} &$0.869$ & $0.967$\\
& \tiny{n\_estimators=$10$ criterion=\texttt{entropy} max\_depth=$50$} &$0.853$ & $0.949$\\
& \tiny{n\_estimators=$10$ criterion=\texttt{entropy} max\_depth=$100$} &$0.852$ & $0.949$\\
& \tiny{n\_estimators=$10$ criterion=\texttt{gini} max\_depth=$50$} &$0.848$ & $0.948$\\
& \tiny{n\_estimators=$10$ criterion=\texttt{gini} max\_depth=$100$} &$0.847$ & $0.948$\\
& \tiny{n\_estimators=$50$ criterion=\texttt{entropy} max\_depth=$10$} &$0.838$ & $0.947$\\
& \tiny{n\_estimators=$100$ criterion=\texttt{entropy} max\_depth=$10$} &$0.838$ & $0.950$\\
& \tiny{n\_estimators=$100$ criterion=\texttt{gini} max\_depth=$10$} &$0.835$ & $0.949$\\
& \tiny{n\_estimators=$50$ criterion=\texttt{gini} max\_depth=$10$} &$0.834$ & $0.945$\\
& \tiny{n\_estimators=$10$ criterion=\texttt{entropy} max\_depth=$10$} &$0.828$ & $0.933$\\
& \tiny{n\_estimators=$10$ criterion=\texttt{gini} max\_depth=$10$} &$0.825$ & $0.930$\\
\midrule
SGDClassifier & \tiny{loss=\texttt{hinge} penalty=\texttt{l2}} &$0.819$ & $0.914$\\
& \tiny{loss=\texttt{perceptron} penalty=\texttt{l1}} &$0.818$ & $0.912$\\
& \tiny{loss=\texttt{modified\_huber} penalty=\texttt{l1}} &$0.817$ & $0.910$\\
& \tiny{loss=\texttt{modified\_huber} penalty=\texttt{l2}} &$0.816$ & $0.913$\\
& \tiny{loss=\texttt{log} penalty=\texttt{elasticnet}} &$0.816$ & $0.912$\\
& \tiny{loss=\texttt{hinge} penalty=\texttt{elasticnet}} &$0.816$ & $0.913$\\
& \tiny{loss=\texttt{squared\_hinge} penalty=\texttt{elasticnet}} &$0.815$ & $0.914$\\
& \tiny{loss=\texttt{hinge} penalty=\texttt{l1}} &$0.815$ & $0.911$\\
& \tiny{loss=\texttt{log} penalty=\texttt{l1}} &$0.815$ & $0.910$\\
& \tiny{loss=\texttt{perceptron} penalty=\texttt{l2}} &$0.814$ & $0.913$\\
& \tiny{loss=\texttt{perceptron} penalty=\texttt{elasticnet}} &$0.814$ & $0.912$\\
& \tiny{loss=\texttt{squared\_hinge} penalty=\texttt{l2}} &$0.814$ & $0.912$\\
& \tiny{loss=\texttt{modified\_huber} penalty=\texttt{elasticnet}} &$0.813$ & $0.914$\\
& \tiny{loss=\texttt{log} penalty=\texttt{l2}} &$0.813$ & $0.913$\\
& \tiny{loss=\texttt{squared\_hinge} penalty=\texttt{l1}} &$0.813$ & $0.911$\\
\midrule
SVC & \tiny{C=$10$ kernel=\texttt{rbf}} &$0.897$ & $0.973$\\
& \tiny{C=$10$ kernel=\texttt{poly}} &$0.891$ & $0.976$\\
& \tiny{C=$100$ kernel=\texttt{poly}} &$0.890$ & $0.978$\\
& \tiny{C=$100$ kernel=\texttt{rbf}} &$0.890$ & $0.972$\\
& \tiny{C=$1$ kernel=\texttt{rbf}} &$0.879$ & $0.966$\\
& \tiny{C=$1$ kernel=\texttt{poly}} &$0.873$ & $0.957$\\
& \tiny{C=$1$ kernel=\texttt{linear}} &$0.839$ & $0.929$\\
& \tiny{C=$10$ kernel=\texttt{linear}} &$0.829$ & $0.927$\\
& \tiny{C=$100$ kernel=\texttt{linear}} &$0.827$ & $0.926$\\
& \tiny{C=$1$ kernel=\texttt{sigmoid}} &$0.678$ & $0.898$\\
& \tiny{C=$10$ kernel=\texttt{sigmoid}} &$0.671$ & $0.873$\\
& \tiny{C=$100$ kernel=\texttt{sigmoid}} &$0.664$ & $0.868$\\
\end{longtable}

\section{Conclusions}

This paper introduced \fmnist, a fashion product images dataset intended to be a drop-in replacement of MNIST and whilst providing a more challenging alternative for benchmarking machine learning algorithm. The images in \fmnist are converted to a format that matches that of the MNIST dataset, making it immediately compatible with any machine learning package capable of working with the original MNIST dataset.

\bibliographystyle{abbrvnat}
\bibliography{ms}

\begin{thebibliography}{6}
\providecommand{\natexlab}[1]{#1}
\providecommand{\url}[1]{\texttt{#1}}
\expandafter\ifx\csname urlstyle\endcsname\relax
  \providecommand{\doi}[1]{doi: #1}\else
  \providecommand{\doi}{doi: \begingroup \urlstyle{rm}\Url}\fi

\bibitem[Ciregan et~al.(2012)Ciregan, Meier, and Schmidhuber]{ciregan2012multi}
D.~Ciregan, U.~Meier, and J.~Schmidhuber.
\newblock Multi-column deep neural networks for image classification.
\newblock In \emph{Computer Vision and Pattern Recognition (CVPR), 2012 IEEE
  Conference on}, pages 3642--3649. IEEE, 2012.

\bibitem[Cohen et~al.(2017)Cohen, Afshar, Tapson, and van
  Schaik]{cohen2017emnist}
G.~Cohen, S.~Afshar, J.~Tapson, and A.~van Schaik.
\newblock Emnist: an extension of mnist to handwritten letters.
\newblock \emph{arXiv preprint arXiv:1702.05373}, 2017.

\bibitem[Deng et~al.(2009)Deng, Dong, Socher, Li, Li, and
  Fei-Fei]{deng2009imagenet}
J.~Deng, W.~Dong, R.~Socher, L.-J. Li, K.~Li, and L.~Fei-Fei.
\newblock Imagenet: A large-scale hierarchical image database.
\newblock In \emph{Computer Vision and Pattern Recognition, 2009. CVPR 2009.
  IEEE Conference on}, pages 248--255. IEEE, 2009.

\bibitem[Krizhevsky and Hinton(2009)]{krizhevsky2009learning}
A.~Krizhevsky and G.~Hinton.
\newblock Learning multiple layers of features from tiny images.
\newblock 2009.

\bibitem[LeCun et~al.(1998)LeCun, Bottou, Bengio, and
  Haffner]{lecun1998gradient}
Y.~LeCun, L.~Bottou, Y.~Bengio, and P.~Haffner.
\newblock Gradient-based learning applied to document recognition.
\newblock \emph{Proceedings of the IEEE}, 86\penalty0 (11):\penalty0
  2278--2324, 1998.

\bibitem[Wan et~al.(2013)Wan, Zeiler, Zhang, Cun, and
  Fergus]{wan2013regularization}
L.~Wan, M.~Zeiler, S.~Zhang, Y.~L. Cun, and R.~Fergus.
\newblock Regularization of neural networks using dropconnect.
\newblock In \emph{Proceedings of the 30th international conference on machine
  learning (ICML-13)}, pages 1058--1066, 2013.

\end{thebibliography}

\end{document}